# Tailoring AI-Driven Reading Scaffolds to the Distinct Needs of Neurodiverse Learners

Soufiane Jhilal[1,2,3] [0000-0001-9403-4129], Eleonora Pasqua[4,5] [0000−0002−7153−6094], Caterina Marchesi[4], Riccardo Corradi[6], and Martina Galletti[1,5] [0009−0002−2079−8999]

[1] Sony Computer Sciences Laboratories, Paris, France
[2] Institut de l'Audition - Institut Pasteur, Paris, France
[3] Université Paris Cité, France
[4] Centro Ricerca e Cura di Roma, Italy
[5] Sapienza Università di Roma, Italy
[6] I-Lab, Rome, Italy
soufiane.jhilal@pasteur.fr
e.pasqua@crc-balbuzie.it
c.marchesi@crc-balbuzie.it
r.corradi@Ilabroma.com
martina.galletti@sony.com

**Abstract.** Neurodiverse learners often require reading supports, yet increasing scaffold richness can sometimes overload attention and working memory rather than improve comprehension. Grounded in the Construction-Integration model and a contingent scaffolding perspective, we examine how structural versus semantic scaffolds shape comprehension and reading experience in a supervised inclusive context. Using an adapted reading interface, we compared four modalities: unmodified text, sentence-segmented text, segmented text with pictograms, and segmented text with pictograms plus keyword labels. In a within-subject pilot with 14 primary-school learners with special educational needs and disabilities, we measured reading comprehension using standardized questions and collected brief child- and therapist-reported experience measures alongside open-ended feedback. Results highlight heterogeneous responses as some learners showed patterns consistent with benefits from segmentation and pictograms, while others showed patterns consistent with increased coordination costs when visual scaffolds were introduced. Experience ratings showed limited differences between modalities, with some apparent effects linked to clinical complexity, particularly for perceived ease of understanding. Open-ended feedback of the learners frequently requested simpler wording and additional visual supports. These findings suggest that no single scaffold is universally optimal, reinforcing the need for calibrated, adjustable scaffolding and provide design implications for human–AI co-regulation in supervised inclusive reading contexts.

**Keywords:** AI for Special Education, Neurodiversity, Multimodal Scaffolding, Reading Comprehension, Inclusive Learning Technologies.

# 1 Introduction

Reading comprehension is a complex cognitive process that involves integrating information from the text, existing knowledge, and contextual cues to construct meaning [1]. This skill develops gradually, requiring learners to decode words, expand vocabulary, and synthesize textual ideas [2, 3]. For neurodiverse children with Special Educational Needs and Disabilities (SEND), such as learning disabilities, developmental disorders, and communication impairments, mastering reading comprehension poses additional challenges. These difficulties often originate from deficits in auditory and visual processing, language skills, attention, and working memory [4]. As a result, many learners experience persistent barriers to independent reading and to classroom participation when text is the dominant medium for instruction and assessment.

A long-standing response to these barriers is individualized intervention delivered by specialized professionals (e.g., speech-language therapy, targeted reading rehabilitation). While effective, this approach is resource-intensive and difficult to scale, especially where specialist availability is limited [5, 6]. This has motivated growing interest in technology-enhanced reading scaffolds that aim to make texts more accessible and provide additional cues for comprehension [7]. However, adding support is not inherently beneficial. In neurodiverse populations in particular, a scaffold that helps one learner may overload another by increasing sensory input, splitting attention, or introducing redundant information that competes with the primary reading task. The central challenge is therefore not simply to provide more supports, but to calibrate the type and intensity of support to the learner and the moment [8].

In this paper, we investigate how different scaffolds shape comprehension and reading experience for neurodiverse learners in a supervised reading context. We use an adapted version of the ARTIS reading interface [9–11] to present different forms of reading support, ranging from structural simplification through sentence segmentation to semantic scaffolds such as pictograms and pictogram-label pairs. In a within-subject pilot study with 14 primary school children with SEND, we combine comprehension outcomes with learner- and therapist-reported experience measures and feedback to characterize when particular scaffolds appear supportive versus distracting, and to derive design implications for calibrated, adjustable AI-enabled reading support in inclusive settings. The following sections review related work and the theoretical grounding that motivates these comparisons.

# 2 Related Work

AI has expanded what is possible in both educational [12] and clinical contexts [13, 14]. Yet comparatively less work has focused on AI-enabled supports that directly target language and reading comprehension for neurodiverse learners with SEND in supervised educational or rehabilitative practice.

Digital platforms increasingly use AI to support reading through features like summarization, question-answering, simplification, and semantic annotation. Earlier systems such as 3D Readers [15] and CACSR [16] illustrate valuable techniques (e.g., main-idea identification, self-questioning, inference generation) but were developed primarily for mainstream educational settings rather than heterogeneous rehabilitation-oriented contexts. RIDInet [17] platform represents a specialized effort in Italian tele-rehabilitation, providing inferential comprehension exercises but lacking multimodal augmentation. Overall, these approaches fail to address the multimodal needs of SEND learners, and to provide a framework suitable for clinical or scalable intervention.

A second strand of related work concerns semantic scaffolds that make key content easier to locate and interpret during reading. One common approach is to highlight keywords to direct attention to core concepts. Keyword cues have a long history in educational psychology [18, 19] and can now be automated with deep learning approaches [20]. However, for many SEND learners, linguistic cues alone may not fully address comprehension barriers. Research shows that students with dyslexia and language disorders often rely on non-verbal and visual-spatial abilities to compensate for linguistic difficulties [21–23], underscoring the value of multimodal interventions that provide visual support alongside textual information. Pictograms, which are simplified visual representations, are known to improve information processing, particularly in learners with language disorders, due to their concrete and universally interpretable nature [24]. On the technical side, text-to-pictogram translation has progressed from rule-based pipelines [25] to systems that incorporate data-driven neural approaches [26]. Despite this progress, relatively few studies examine keyword cues and pictogram scaffolds together as part of the reading experience for neurodiverse learners, or test whether increasing scaffold richness can backfire.

## 3 Theoretical Framework

Our study is grounded in the idea that scaffolds can reduce some demands while increasing others, so effects need not be monotonic. We adopt the Kintsch and van Dijk's Construction-Integration account of comprehension [27, 28], which theorizes that discourse comprehension proceeds in two main phases: (1) the construction phase, in which readers activate a network of propositions derived from textual input and prior knowledge, and (2) the integration phase, in which this network is refined into a coherent mental representation by resolving inconsistencies and strengthening relevant connections. This maps onto our scaffold mechanisms as sentence segmentation supports early processing by lowering linguistic density and enabling incremental construction, while pictograms and pictogram-label pairs target integration by grounding key concepts and making symbol-word mappings more explicit.

We also draw on a scaffolding perspective in which support is beneficial when it is contingent and aligned with the learner's current capabilities. In Vygotskian terms, assistance is most productive when it operates within the learner's zone of proximal development and supports gradual autonomy [29]. In supervised inclusive settings, such alignment is often actively managed by teachers and clinicians, who adjust task demands and supports in response to attention, fatigue, and comprehension difficulties.

Finally, we conceptualize multimodal scaffolds as a trade-off between grounding benefits and coordination costs. Visual symbols can reduce reliance on fragile decoding or lexical access, but they can also increase attentional switching and working-memory demands when learners must coordinate multiple representations or process redundant cues. This trade-off is particularly salient for neurodiverse learners for whom attentional and executive resources may be limited and fluctuate across time and tasks [4]. Accordingly, we interpret outcomes in terms of this balance and emphasize within-learner comparisons, rather than assuming average gains from adding features.

## 4 Methods

### 4.1 User Interface and Multimodal Presentation

ARTIS is an AI-powered reading support platform designed for supervised educational and rehabilitation sessions with neurodiverse learners, where a therapist or teacher can adjust scaffolding levels to the learner and session goals [9–11]. The text-to-pictogram pipeline used in this study is an adapted version of the original ARTIS workflow. The updated pipeline [30] is multilingual and context-aware. It first detects the input language and segments the text into sentences, then extracts the top candidate keywords per sentence using YAKE [31], prioritizing semantic salience. Each keyword is queried against ARASAAC [32], retrieving pictogram candidates accompanied by a semantic definition. A disambiguation module then selects the most contextually appropriate pictogram by matching sentence-context embeddings to pictogram definitions via cosine similarity using multilingual sentence transformers [33], reducing polysemy errors without manual intervention. Selected pictograms are then ordered to follow the original sentence structure and integrated into the interface so that pictograms align with the corresponding textual content. Prior to the study, clinicians performed a targeted review of representative outputs to ensure developmental and therapeutic appropriateness.

To examine progressively richer scaffolds, text was presented through four distinct modalities:

- **Modality A** serves as a baseline and shows the unmodified full text;
- **Modality B** segments the text into sentences for improved readability and reduced cognitive load;
- **Modality C** adds pictograms aligned to each segmented sentence, providing visual cues that aid semantic interpretation; and
- **Modality D** adds keyword labels beneath the pictograms, allowing for explicit associations between visual symbols and their corresponding textual terms.

These modalities are based on the Construction-Integration Model in which readers construct candidate propositions from text and integrate them into a coherent situation model by strengthening relevant links and resolving inconsistencies. Each modality was crafted to support these phases by reducing cognitive load and enhancing the formation of coherent propositional structures. By progressively augmenting the text from Modality A to Modality D with segmentation, imagery, and explicit labeling, the interface aligns with the Construction-Integration Model's emphasis on the interaction between surface structures, text base representations, and situation models, thereby scaffolding both micro-level sentence processing and macro-level discourse comprehension. Figure 1 illustrates the four modalities as presented in the interface.

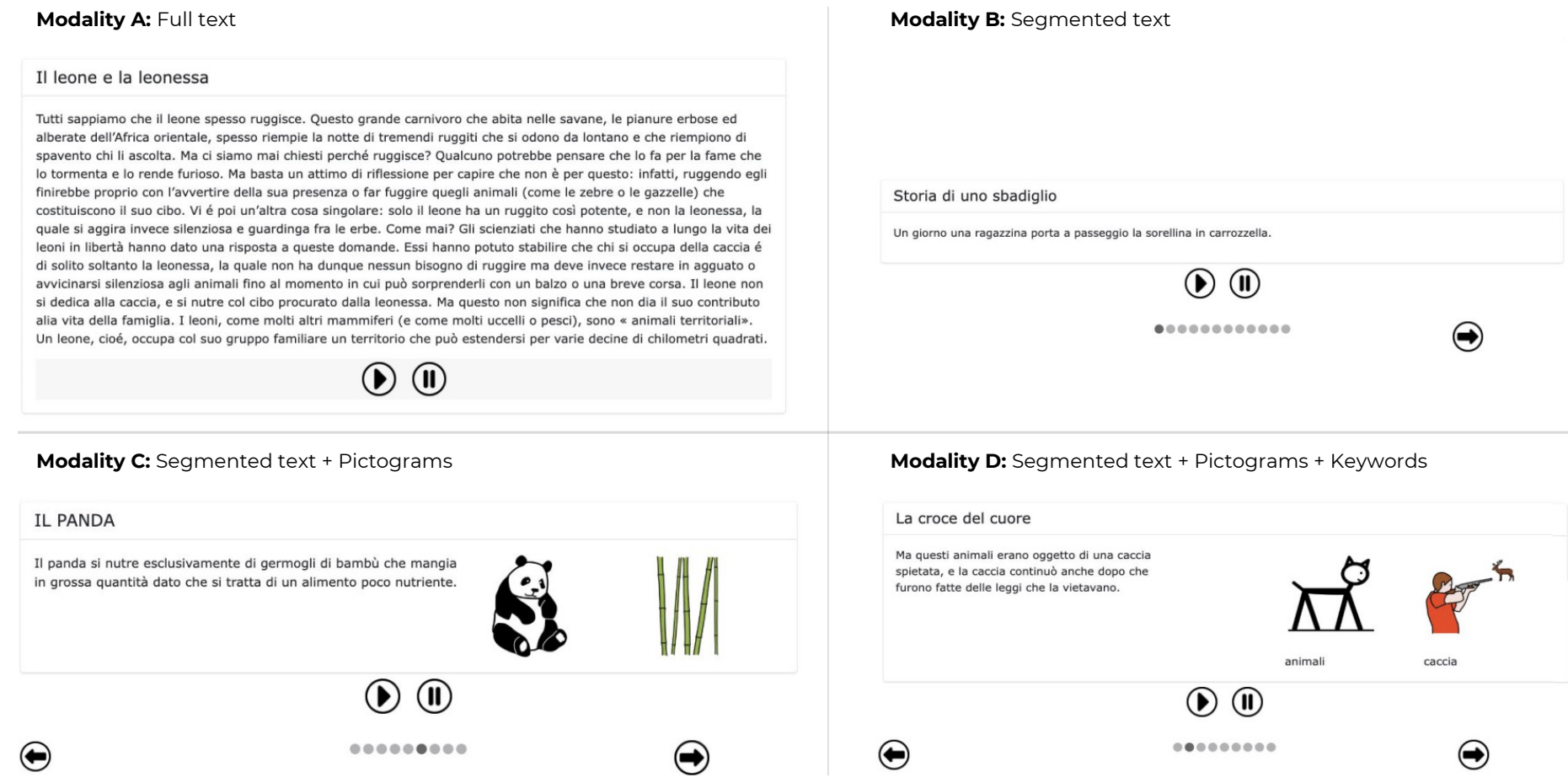

**Fig. 1.** Interface screenshots showing the four reading modalities: plain text (A), sentence segmentation (B), segmentation with pictograms (C), and segmentation with pictograms and keyword labels (D).

## 4.2 Experimental Design

This exploratory pilot study examined how different scaffold types and levels of richness shape both reading comprehension and the reading experience for neurodiverse learners with SEND in a supervised context. We compared four interface modalities that progressively vary the form and intensity of support (Modality A–D), from unmodified text to sentence segmentation and symbol-based semantic cues (pictograms, with or without keyword labels). While this study focuses on the effects of these scaffolds on neurodiverse learners, a separate technical evaluation of the pipeline's multilingual performance, pictogram coverage, and semantic accuracy has also been conducted [30].

We used a within-subject design in which each participant experienced all four modalities. Passages were drawn from the standardized MT reading comprehension test bank [34], selected to match the participant's school grade. These materials were chosen for their well-established use in both didactic and clinical settings, and because they are standardized for primary education in Italy. Eight passages from the MT test bank were used in total, four per grade level (4th and 5th grade), each comprising 10–14 multiple-choice comprehension questions. Each participant completed one passage per modality, such that every modality was paired with a different text. To mitigate order and text effects, the assignment of passages to modalities was varied across participants, ensuring that no single text was consistently paired with the same modality. After each passage, participants answered the comprehension questions, with performance operationalized as the percentage of correct responses per modality.

To capture the lived experience of reading beyond accuracy alone, we collected brief post-condition questionnaires from both learners and supervising therapists. Learners reported perceived ease of reading and understanding, engagement, and concentration, and provided open-ended feedback on what would make the experience easier or more enjoyable. Therapists reported observations of motivation, attention, frustration, fatigue, and perceived usefulness of the support, alongside open-ended comments about difficulties encountered and any support or adjustments provided during the session. To minimize burden and avoid confusion for young participants, we used a short, age-appropriate learner questionnaire with a 3-point Likert scale (1 = Disagree, 2 = Neutral, 3 = Agree), while therapists completed a parallel set of items on a 5-point Likert scale (1 = Strongly disagree to 5 = Strongly agree) to allow for finer-grained clinical judgment. The learner questionnaire included only a small number of items to reduce fatigue and maintain attention. During completion, the supervising therapist guided the child through each question (e.g., clarifying the meaning and checking understanding), discussed how the child felt about the just-completed modality, and then recorded the child's response together with them. Table 1 summarizes all questionnaire items administered to learners and therapists.

**Table 1.** Questionnaire items for children and therapists. Children answered four Likert items on a 3-point scale (1–3) plus one open-ended question; therapists answered seven Likert items on a 5-point scale (1–5) plus two conditional open-ended follow-ups.

| **Children questions** | |
|---|---|
| Question 1 - Likert (1–3) | The text was easy to read. |
| Question 2 - Likert (1–3) | The text was easy to understand. |
| Question 3 - Likert (1–3) | I was engaged while reading. |
| Question 4 - Likert (1–3) | I was focused while reading. |
| Question 5 - Open-ended | What would you change to make reading even easier or more fun in this mode? |
| **Therapist questions** | |
| Question 1 - Likert (1–5) | The child seems comfortable while reading. |
| Question 2 - Likert (1–5) | The reading interface worked effectively. |
| Question 3 - Open-ended | If you selected strongly disagree or disagree, specify what went wrong. |
| Question 4 - Likert (1–5) | The child is attentive while reading. |
| Question 5 - Likert (1–5) | The child was motivated in reading. |
| Question 6 - Likert (1–5) | The child requested support from the operator. |
| Question 7 - Open-ended | If you selected agree or strongly agree, specify what type of support was requested. |
| Question 8 - Likert (1–5) | The child showed signs of fatigue in performing the activity. |
| Question 9 - Likert (1–5) | The child showed signs of frustration in reading. |

### 4.3 Derived Measures of Comprehension Gains

In addition to raw comprehension scores, we computed two derived metrics to assess relative improvement (per participant) in performance across the scaffolded interface modalities. These metrics do not rely on the actual order in which modalities were presented to participants but instead reflect the logical progression of support from Modality A to Modality D.

**Absolute Modality Gain** represents the relative improvement in comprehension score for each modality compared to the participant's baseline in Modality A.

$$AbsoluteGain = \frac{Score_{current\ modality} - Score_{modality\ A}}{Score_{modality\ A}}$$

**Relative Modality Gain** measures the incremental improvement relative to the preceding modality, reflecting how each added support feature (sentence segmentation, pictograms, keyword labels) impacted comprehension.

$$RelativeGain = \frac{Score_{current\ modality} - Score_{prior\ modality}}{Score_{prior\ modality}}$$

These gain metrics normalize performance relative to each participant's own baseline or to the previous scaffolded step. They offer a more individualized assessment of how each added support feature contributes to comprehension, which is particularly valuable in small-sample, within-subject designs with heterogeneous cognitive profiles. In our dataset, comprehension scores were non-zero across all participants and modalities, so the ratio-based gain metrics were well-defined for all observations.

### 4.4 Participants

Fourteen neurodiverse primary-school children participated in this exploratory study. Participants were recruited through CRC (Centro Ricerca e Cura), a developmental rehabilitation center in Rome, Italy, where the study was conducted as part of supervised reading sessions. Participation occurred with informed consent from guardians and assent from children, and data were handled in a de-identified manner. The sample size (N = 14) and heterogeneity of profiles reflect the study's goal of generating early design evidence about how different scaffold types may help or overload different learners, rather than estimating population-level effects.

At the time of testing, children were 9–11 years old (M = 10.6), with 10 males and 4 females, and were enrolled in either 4th or 5th grade. All participants had at least one clinically established neurodevelopmental and/or learning-related diagnosis assigned by qualified professionals according to local clinical practice. These included:

- **Developmental Coordination Disorder (DCD)** characterized by impaired motor coordination that interferes with academic or daily functioning;
- **Attention-Deficit/Hyperactivity Disorder (ADHD)** marked by persistent inattention, hyperactivity, and/or impulsivity;
- **Language Disorder (LD)** involving significant difficulties in understanding or producing spoken language;
- **Specific Learning Disorder (SLD)** encompassing difficulties in reading, writing, or arithmetic (e.g., dyslexia, dysorthographia);
- **Mixed Developmental Disorder (MMD)** referring to combined delays in multiple developmental domains; and

- **Emotional Disorder (ED)** involving challenges in emotional regulation, such as heightened anxiety or mood related symptoms. Table 2 summarizes participant characteristics and diagnostic profiles.

**Table 2.** Participant profiles including demographic information and diagnosed conditions. Grade refers to school grade (4 = 4th grade, 5 = 5th grade). *X indicates the presence of a diagnosis; – indicates absence.*

| ID | Gender | Age | Grade | DCD | ADHD | LD | SLD | MMD | ED |
|---|---|---|---|---|---|---|---|---|---|
| user1 | Male | 10 | 4 | – | X | X | – | – | – |
| user2 | Male | 10 | 4 | – | X | – | X | – | – |
| user3 | Male | 11 | 5 | – | X | – | X | – | – |
| user4 | Male | 10 | 5 | – | – | X | X | – | X |
| user5 | Female | 10 | 4 | – | – | – | X | – | – |
| user6 | Male | 10 | 4 | X | X | X | – | – | – |
| user7 | Male | 10 | 5 | – | X | – | X | – | – |
| user8 | Male | 11 | 4 | – | – | – | – | – | X |
| user9 | Female | 9 | 4 | – | – | – | – | X | – |
| user10 | Male | 11 | 5 | X | – | X | X | – | – |
| user11 | Male | 11 | 5 | – | X | – | X | – | – |
| user12 | Female | 10 | 5 | – | – | – | X | – | – |
| user13 | Female | 11 | 4 | X | X | X | – | – | – |
| user14 | Male | 10 | 4 | – | – | – | X | – | X |

**Clinical Complexity and Comorbidity**. To better understand how individual clinical profiles influenced reading comprehension performance, we computed a comorbidity index for each participant, representing the total number of diagnosed neurodevelopmental or learning related conditions. This allowed us to assess how the degree of clinical complexity related to comprehension outcomes and different types of scaffolded support.

## 5 Results

We used mixed-design ANOVAs with Modality (A–D) as a within-participant factor and diagnostic indicator tested separately (presence vs. absence of each diagnosis) or comorbidity level as between-participant factors, to identify potentially meaningful patterns in scaffold response. Given the pilot scale (N = 14), these analyses are intended to highlight indicative trends that can inform interpretation and future hypothesis testing rather than to provide definitive population-level estimates. Prior to analysis, standard assumption checks were conducted for residual normality and homogeneity of variance. Post-hoc comparisons were corrected for multiple testing using a Bonferroni adjustment.

### 5.1 Diagnostic Group Differences in Performance

Mixed-design ANOVAs examining the effects of Modality and diagnosis on comprehension scores suggested systematic performance differences linked to several diagnoses (Fig. 2). In this sample, children with Developmental Coordination Disorder (DCD) scored lower than peers without DCD ($F(1,48) = 6.15$, $p = .017$), as did those with Language Disorders (LD) ($F(1,48) = 6.49$, $p = .014$), while learners with Emotional Disorders (ED) showed higher overall scores ($F(1,48) = 8.38$, $p = .006$). Children with DCD and LD tended to underperform across modalities, with an apparent increase in scores in Modality C. Conversely, children with ED performed relatively well in Modalities A and B (plain and segmented text), but their scores were lower in Modality C.

### 5.2 Diagnostic Modulation of Multimodal Benefit

To examine whether the relative benefit of multimodal scaffolds varied across diagnostic profiles, we analyzed both absolute gain from Modality A (Fig. 3) and relative gain between successive scaffolding steps (Fig. 4).

Relative Modality Gain showed interaction effects between Modality and diagnosis for ED ($F(3,48) = 4.18$, $p = .010$) and Language Disorders (LD) ($F(3,48) = 7.97$, $p < .001$), consistent with the possibility that scaffold effects differed across learner profiles. For ED, relative gain was near zero from Modality A to B (segmented text) but decreased in Modality C, suggesting that pictograms may have introduced additional coordination demands for some learners in this group. In contrast, learners with LD showed larger relative gains in Modalities B and C, consistent with the hypothesis that both sentence segmentation and pictograms can provide useful support in this sample. However, gains for the LD group decreased in Modality D relative to Modality C, suggesting that adding keyword labels may not provide additional benefit for some learners. Absolute Modality Gain also suggested interaction effects for DCD ($F(3,48) = 2.80$, $p = .050$) and a trend for LD ($F(3,48) = 2.39$, $p = .080$).

Overall, these patterns point to heterogeneous scaffold responses across modalities and learner profiles within this pilot sample.

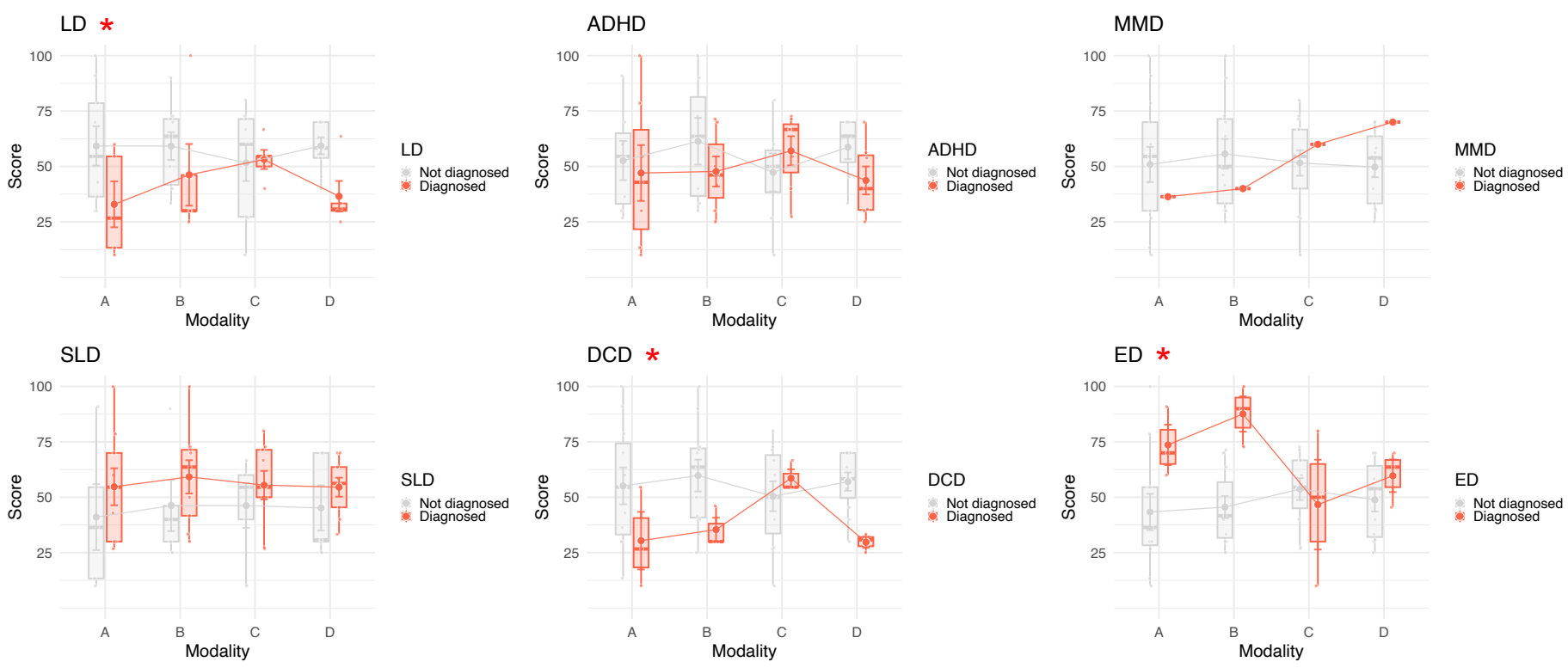

**Fig. 2.** Comprehension scores across modalities by diagnosis. *Red asterisk* (*) = *significant main effect of diagnosis.*

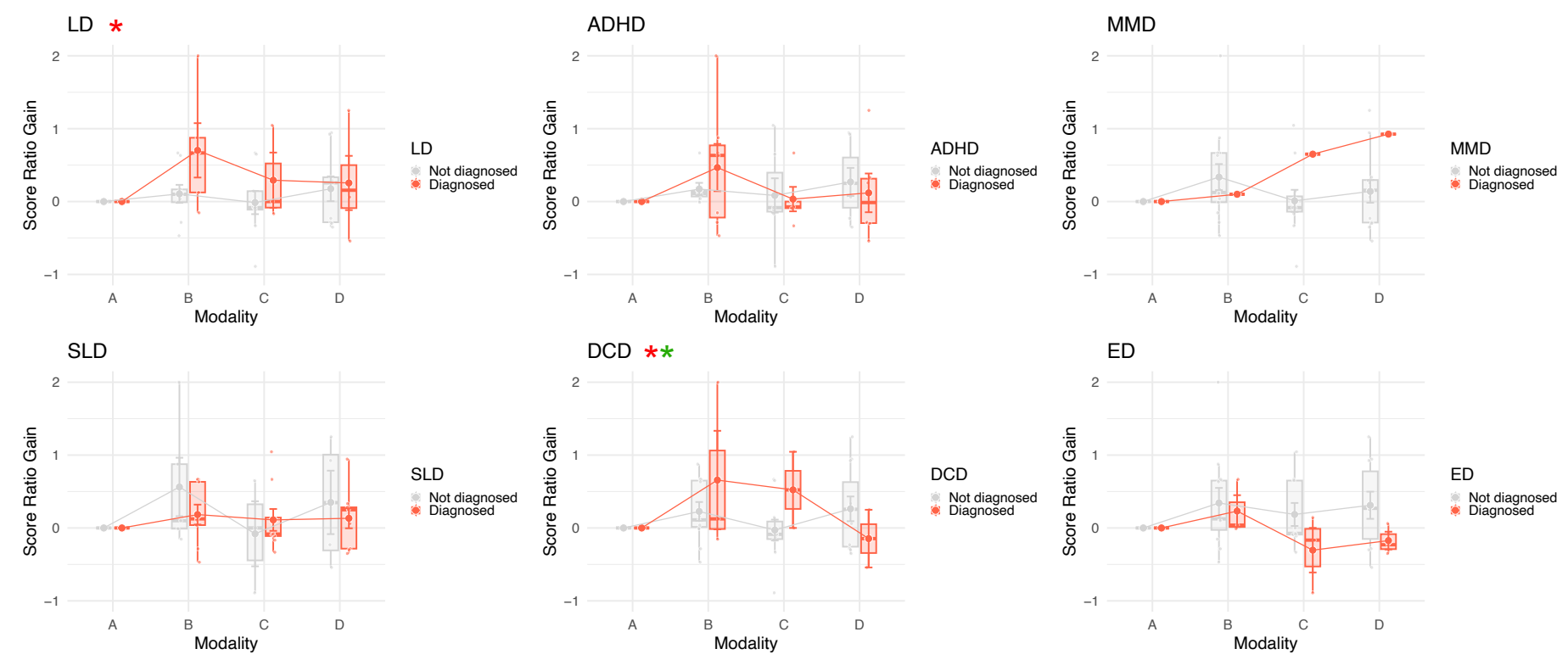

**Fig. 3.** Absolute Modality Gain by diagnosis. *Red asterisk* (*) *= significant main effect of diagnosis; Green asterisk* (*) *= significant interaction effect between Modality and diagnosis.*

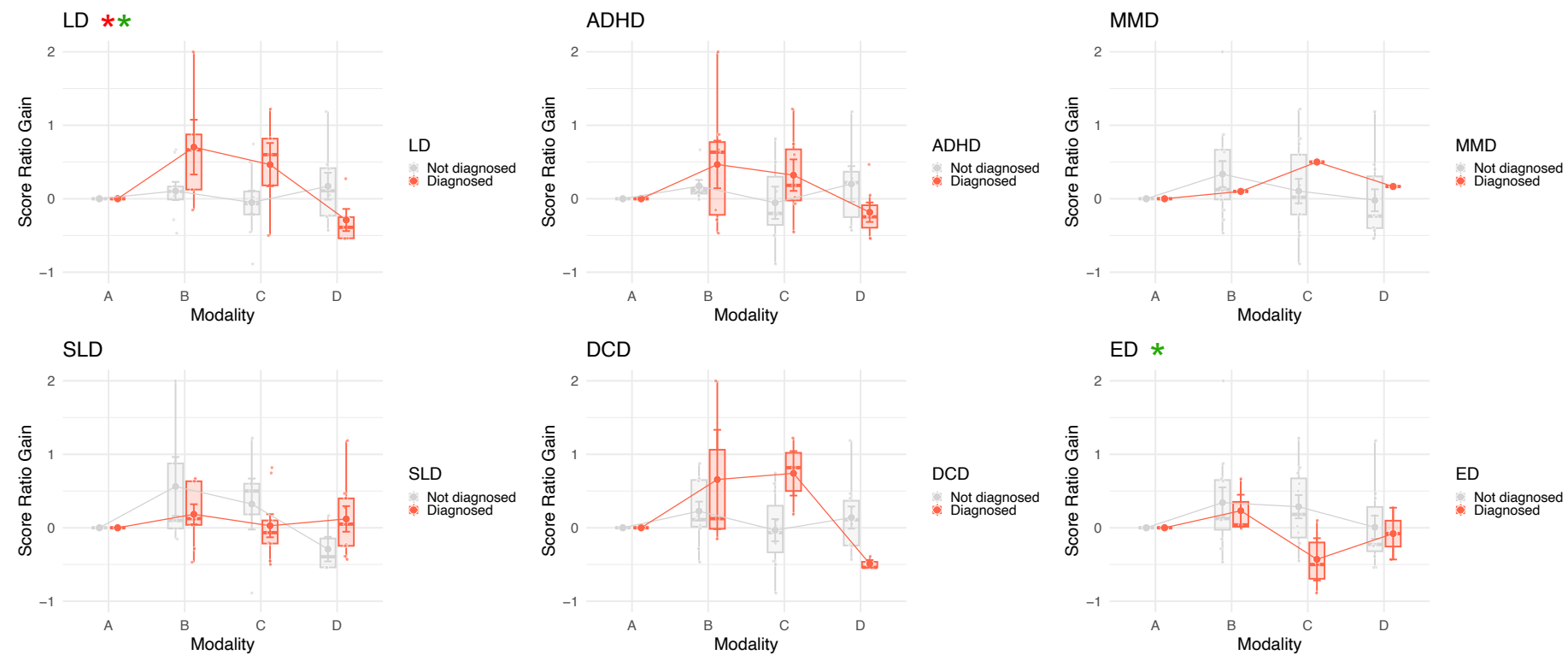

**Fig. 4.** Relative Modality Gain by diagnosis. *Red asterisk* (*) *= significant main effect of diagnosis; Green asterisk* (*) *= significant interaction effect between Modality and diagnosis.*

## 5.3 Impact of Comorbidity

We also examined whether clinical complexity, operationalized as the number of diagnosed conditions, was associated with comprehension or scaffold response. There was no main effect of comorbidity on comprehension scores ($F(2,44) = .81$, $p = .452$). However, an interaction between Modality and comorbidity was observed for Relative Modality Gain ($F(6,44) = 3.03$, $p = .014$), suggesting that learners with different levels of comorbidity showed different patterns of change across scaffold steps (Fig. 5).

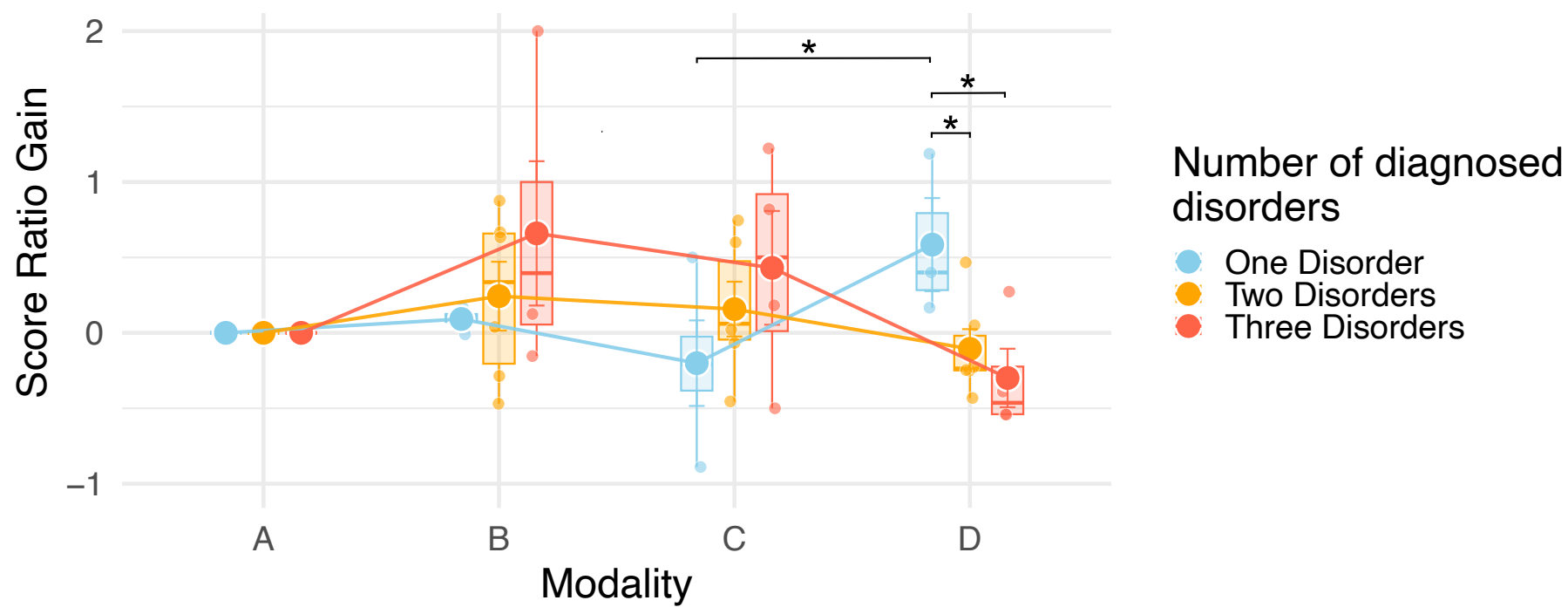

**Fig. 5.** Relative Modality Gain by number of diagnosed disorders across modalities. *Asterisk* (*) *indicates a significant post-hoc comparison.*

### 5.4 Qualitative Insights from Learner Feedback

Across the learner and therapist Likert-scale questionnaires, we observed limited variation in ratings between modalities for the group as a whole. However, a distinct pattern emerged regarding clinical complexity for the children's Question 1 ("*The text was easy to understand*", see Fig. 6). Ratings indicated that children with the highest comorbidity (3 diagnosed conditions) perceived the texts as harder to understand compared to peers with fewer diagnoses. This difficulty appeared most acute in the baseline condition (Modality A). Specifically, the 3-disorder subgroup rated the unmodified text notably lower than the scaffolded conditions (B, C, and D), suggesting that for clinically more complex learners, the baseline presentation was experienced as harder to follow relative to segmented and multimodally scaffolded variants.

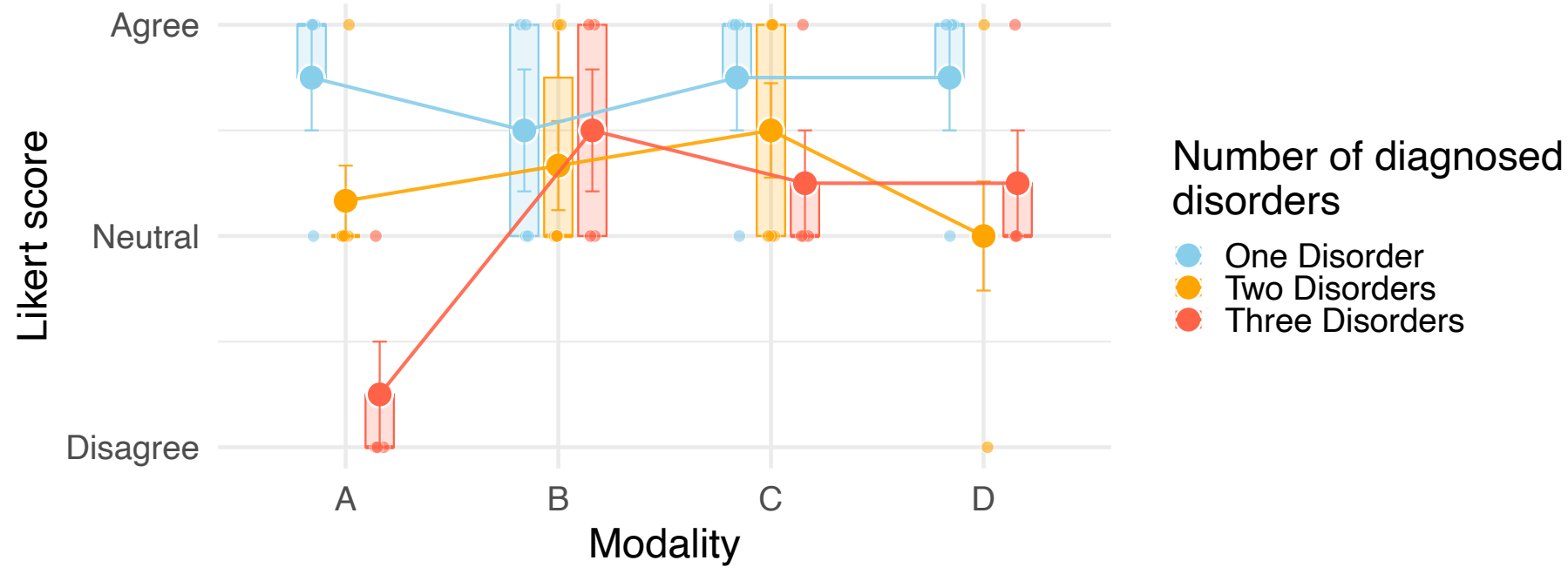

**Fig. 6.** Children's Likert scale ratings for Question 1 "*The text was easy to understand*" by number of diagnosed disorders across modalities.

Complementing these ratings, the open-ended responses provided additional context about perceived barriers and preferences. When asked what would make reading easier or more enjoyable, children most frequently requested simpler words, pointing to the overall difficulty of the texts. The second most common suggestion was to include more drawings or images, consistent with a perceived benefit or preference for visual support in the reading material.

## 6 Discussion

This exploratory pilot examined how progressively richer reading scaffolds (sentence segmentation, pictograms, and pictogram-keyword pairs) relate to comprehension and reading experience in a supervised SEND context. Rather than showing a uniform benefit of adding support, the results point to heterogeneous scaffold responses that vary across diagnostic profiles and clinical complexity, consistent with our framing of a trade-off between grounding benefits and coordination costs.

On comprehension outcomes, several diagnoses were associated with overall performance differences across modalities, indicating that baseline comprehension levels differed by profile in this sample. Beyond these main effects, analyses of gain metrics suggested that the pattern of change across scaffold steps may differ by learner group.

Children with Language Disorders (LD) appear to benefit the most from added scaffolds, especially when text was segmented (Modality B) and augmented with pictograms (Modality C). These results are in line with previous research suggesting that learners with decoding or attentional difficulties gain from structural simplification and visual cues [35, 36]. However, their benefit diminished in Modality D, where keyword labels were added beneath pictograms. This pattern suggests that additional textual information, despite being supportive in principle, may become redundant or even distracting when cognitive or linguistic resources are limited.

Conversely, learners with Emotional Disorders (ED) showed an opposite trend: they performed relatively well in Modality A (raw text) and Modality B (segmented text), but their performance declined in Modality C. The introduction of visual elements may have disrupted their cognitive processing, either due to increased sensory load or reduced ability to integrate multimodal content. This highlights the need to avoid over-scaffolding for some learners, as even well-intentioned supports can become counterproductive when not aligned with individual processing preferences.

Interestingly, learners with Developmental Coordination Disorder (DCD) and Language Disorders (LD), who initially showed lower performance, achieved comparable scores to peers in Modality C. One possible interpretation is that the addition of pictograms without text labels may have helped them bypass linguistic and decoding challenges by offering a more intuitive route to meaning-making. This could mean that visual support can enhance accessibility without overwhelming cognitive resources.

Comorbidity analyses similarly highlighted variability in scaffold response. Although comorbidity level was not associated with overall comprehension scores, it did modulate Relative Modality Gain, indicating that learners with different levels of clinical complexity showed different trajectories across scaffold steps. In the qualitative questionnaires, most Likert items did not show any clear modality differences. The exception was the children's "*text was easy to understand*" item (Fig. 6), which seems to show lower ratings in the 3-disorder subgroup, particularly for the baseline presentation (Modality A). Together, these results suggest that clinical complexity may relate not only to performance but also to perceived ease, and that baseline presentations can be harder to follow for some learners with more complex profiles.

Open-ended feedback provides additional context for these quantitative patterns. Children most frequently requested simpler words, reflecting the general difficulty of the standardized passages, and also commonly requested more drawings or images. While these comments do not establish efficacy, they indicate that visual supports are salient and often perceived as desirable, reinforcing the importance of offering scaffolds that can be adjusted to learner needs and tolerance for added information.

Overall, the findings argue against a one-size-fits-all approach to reading support in heterogeneous SEND populations and highlight several design implications for AI-driven reading interventions. First, scaffolds should dynamically adapt to individual learner profiles rather than relying solely on diagnostic categories, as substantial variability was observed even within groups. Second, systems should carefully balance informational support with cognitive load, as additional visual or semantic cues may either support comprehension or lead to overload depending on the learner. Finally, AI-driven interfaces should allow for adjustable scaffold intensity, enabling therapists or learners themselves to tailor the level and type of support to situational needs. However, these findings should be interpreted in light of the study's scope. As noted in the Participants section, this was an exploratory pilot with a small and clinically diverse sample, intended to explore patterns of differential response across scaffolds rather than produce population-level estimates. This approach enabled in-depth, diagnosis-specific observations that may be overlooked in larger but less targeted studies. Nonetheless, the limited sample size reduces statistical power, and diagnostic categories were not evenly represented. The absence of a neurotypical control group limits comparative interpretation. Therapist supervision, essential for accessibility and engagement, may have influenced task dynamics. Addressing these limitations in subsequent research, will be essential to establishing evidence-based guidelines for personalized reading support in educational technology.

## 7 Conclusion

This exploratory pilot study evaluated four progressively scaffolded reading interface modalities in a supervised SEND context, combining comprehension outcomes with learner- and therapist-reported experience measures. The results suggest that no single support modality is universally effective: optimal scaffolding depends on diagnosis and clinical complexity, with some learners benefiting from rich multimodal reinforcement and others from simpler structural interventions. This variability underscores the importance of dynamic learner modeling in future AI-education systems, enabling real-time adaptation of support strategies to maximize engagement and comprehension. Building on these preliminary insights, future research will scale to larger and more balanced cohorts, explore longitudinal effects, and integrate adaptive algorithms capable of adjusting scaffolding in real time based on continuous learner performance and feedback.

## References

1. Hulme C, Snowling MJ (2013) Learning to Read: What We Know and What We Need to Understand Better. Child Dev Perspect 7:1–5. https://doi.org/10.1111/cdep.12005
2. Scarborough H, Fletcher-Campbell F, Soler J, Reid G (2009) Connecting early language and literacy to later reading (dis)abilities: Evidence, theory, and practice. Approaching difficulties in literacy development: assessment, pedagogy, and programmes 23–39
3. Castles A, Rastle K, Nation K (2018) Ending the Reading Wars: Reading Acquisition From Novice to Expert. Psychol Sci Public Interest 19:5–51. https://doi.org/10.1177/1529100618772271
4. Capin P, Cho E, Miciak J, et al (2021) Examining the Reading and Cognitive Profiles of Students With Significant Reading Comprehension Difficulties. Learning Disability Quarterly 44:183–196. https://doi.org/10.1177/0731948721989973
5. Billingsley BS (2004) Special Education Teacher Retention and Attrition: A Critical Analysis of the Research Literature. J Spec Educ 38:39–55. https://doi.org/10.1177/00224669040380010401
6. Squires K (2013) Addressing the Shortage of Speech-Language Pathologists in School Settings. Journal of the American Academy of Special Education Professionals 131–137
7. Yang Y, Chen L, He W, et al (2025) Artificial Intelligence for Enhancing Special Education for K-12: A Decade of Trends, Themes, and Global Insights (2013–2023). Int J Artif Intell Educ 35:1129–1177. https://doi.org/10.1007/s40593-024-00422-0
8. Cesaroni V, Pasqua E, Bisconti P, Galletti M (2025) A Participatory Strategy for AI Ethics in Education and Rehabilitation Grounded in the Capability Approach. In: Cristea AI, Walker E, Lu Y, et al (eds) Artificial Intelligence in Education. Springer Nature Switzerland, Cham, pp 77–84
9. Galletti M, Pasqua E, Calanca M, et al (2024) ARTIS: a digital interface to promote the rehabiliatation of text comprehension difficulties through Artificial Intelligence
10. Galletti M, Pasqua E, Bianchi F, et al (2023) A Reading Comprehension Interface for Students with Learning Disorders. In: International Conference on Multimodal Interaction. ACM, Paris France, pp 282–287
11. Galletti M (2026) A computational framework for AI-driven reading comprehension rehabilitation for neurodiverse learners. Università degli Studi di Roma "La Sapienza"
12. Wang S, Wang F, Zhu Z, et al (2024) Artificial intelligence in education: A systematic literature review. Expert Systems with Applications 252:124167. https://doi.org/10.1016/j.eswa.2024.124167
13. Zhang Y, Weng Y, Lund J (2022) Applications of Explainable Artificial Intelligence in Diagnosis and Surgery. Diagnostics 12:237. https://doi.org/10.3390/diagnostics12020237
14. Jhilal S, Marchesotti S, Thirion B, et al (2025) Implantable Neural Speech Decoders: Recent Advances, Future Challenges. Neurorehabil Neural Repair. https://doi.org/10.1177/15459683251369468
15. Johnson-Glenberg MC (2007) Web-based reading comprehension instruction: Three studies of 3D-readers. In: Reading comprehension strategies: Theories, interventions, and technologies. Lawrence Erlbaum Associates Publishers, Mahwah, NJ, US, pp 293–324
16. Kim A-H, Vaughn S, Klingner JK, et al (2006) Improving the Reading Comprehension of Middle School Students With Disabilities Through Computer-Assisted Collaborative Strategic Reading. Remedial and Special Education 27:235–249

17. RIDInet. https://www.anastasis.it/ridinet/. Accessed 2 Feb 2026

18. McDaniel MA, Pressley M (1989) Keyword and context instruction of new vocabulary meanings: Effects on text comprehension and memory. Journal of Educational Psychology 81:204–213. https://doi.org/10.1037/0022-0663.81.2.204

19. Seki Y, Akahori K, Sakamoto T (1993) Using Key Words to Facilitate Text Comprehension. Educational technology research 16:11–21

20. Nadim M, Akopian D, Matamoros A (2023) A Comparative Assessment of Unsupervised Keyword Extraction Tools. IEEE Access 11:144778–144798

21. Danis E, Nader A-M, Degré-Pelletier J, Soulières I (2023) Semantic and Visuospatial Fluid Reasoning in School-Aged Autistic Children. J Autism Dev Disord 53:4719–4730. https://doi.org/10.1007/s10803-022-05746-1

22. Jhilal S, Molinaro N, Klimovich-Gray A (2025) Non-verbal skills in auditory word processing: implications for typical and dyslexic readers. Language, Cognition and Neuroscience 40:341–359. https://doi.org/10.1080/23273798.2024.2438012

23. Superbia-Guimarães L, Bader M, Camos V (2023) Can children and adolescents with ADHD use attention to maintain verbal information in working memory? PLOS ONE 18:e0282896. https://doi.org/10.1371/journal.pone.0282896

24. Vaschalde C, Trial P, Esperança-Rodier E, et al (2018) Automatic pictogram generation from speech to help the implementation of a mediated communication. In: Conference on Barrier-free Communication. Geneva, Switzerland

25. Vandeghinste V, Sevens ISL, Eynde FV (2015) Translating text into pictographs. Natural Language Engineering 23:217–244. https://doi.org/10.1017/S135132491500039X

26. Pereira JA, Macêdo D, Zanchettin C, et al (2022) PictoBERT: Transformers for next pictogram prediction. Expert Systems with Applications 202:117231

27. Kintsch W (1988) The role of knowledge in discourse comprehension: A construction-integration model. Psychological Review 95:163–182. https://doi.org/10.1037/0033-295X.95.2.163

28. Kintsch W, van Dijk TA (1978) Toward a model of text comprehension and production. Psychological Review 85:363–394. https://doi.org/10.1037/0033-295X.85.5.363

29. Vygotsky LS (1978) Mind in Society: Development of Higher Psychological Processes. Harvard University Press

30. Jhilal S, Galletti M (2026) Robust Multilingual Text-to-Pictogram Mapping for Scalable Reading Rehabilitation. https://doi.org/10.48550/arXiv.2603.24536

31. YAKE. https://github.com/INESCTEC/yake. Accessed 5 Feb 2026

32. ARASAAC. https://beta.arasaac.org/. Accessed 2 Feb 2026

33. SentenceTransformers. https://sbert.net/. Accessed 5 Feb 2026

34. Cornoldi C, Colpo G, Carretti B (1998) Nuove prove di lettura MT per la scuola media inferiore. Giunti EDU, Firenze

35. Lorch Jr. RF, Lorch EP (1995) Effects of organizational signals on text-processing strategies. Journal of Educational Psychology 87:537–544. https://doi.org/10.1037/0022-0663.87.4.537

36. Chun DM, Plass JL (1997) Research on text comprehension in multimedia environments. https://doi.org/10.64152/10125/25004